\documentclass[conference]{IEEEtran}
\IEEEoverridecommandlockouts
\usepackage{cite}
\usepackage{amsmath,amssymb,amsfonts}
\usepackage{algorithmic}
\usepackage{graphicx}
\usepackage{textcomp}
\usepackage{xcolor}
\usepackage{textalpha} 

\usepackage{hyperref} 
\usepackage{multirow}
\usepackage{booktabs}
\usepackage{siunitx}
\def\BibTeX{{\rm B\kern-.05em{\sc i\kern-.025em b}\kern-.08em
    T\kern-.1667em\lower.7ex\hbox{E}\kern-.125emX}}
\begin{document}

\title{Frailty Estimation in Elderly Oncology Patients Using Multimodal Wearable Data and Multi-Instance Learning\\
}

\author{
\IEEEauthorblockN{Ioannis Kyprakis}
\IEEEauthorblockA{\textit{ECE Dept. HMU}\\ 
\textit{FORTH-ICS}\\
Heraklion, Greece\\
ikyprakis@ics.forth.gr}
\and
\IEEEauthorblockN{Vasileios Skaramagkas}
\IEEEauthorblockA{\textit{ECE Dept. HMU}\\ 
\textit{FORTH-ICS}\\
Heraklion, Greece\\
vskaramag@ics.forth.gr}
\and
\IEEEauthorblockN{Georgia Karanasiou}
\IEEEauthorblockA{\textit{FORTH-BRI}\\
Ioannina, Greece\\
g.karanasiou@gmail.com}
\and
\IEEEauthorblockN{Lampros Lakkas}
\IEEEauthorblockA{\textit{Univ. of Ioannina}\\
\textit{(Med.) Physiology Lab}\\
Ioannina, Greece\\
l.lakkas@uoi.gr}
\and
\IEEEauthorblockN{Andri Papakonstantinou}
\IEEEauthorblockA{\textit{Breast Center}\\
\textit{Karolinska Univ. Hosp.}\\
Stockholm, Sweden\\
andri.papakonstantinou@ki.se}
\and
\IEEEauthorblockN{Domen Ribnikar}
\IEEEauthorblockA{\textit{Inst. of Oncology Ljubljana}\\
Ljubljana, Slovenia\\
dribnikar@onko-i.si}
\and
\IEEEauthorblockN{Kalliopi Keramida}
\IEEEauthorblockA{\textit{NKUA Med. Sch.; Attikon}\\
\textit{Hosp., Agios Savvas Hosp.}\\
Athens, Greece\\
keramidakalliopi@hotmail.com}
\and
\IEEEauthorblockN{Dorothea Tsekoura}
\IEEEauthorblockA{\textit{Aretaieio Univ. Hosp.,}\\
NKUA, Athens, Greece\\
dtsekoura@med.uoa.gr}
\and
\IEEEauthorblockN{Ketti Mazzocco}
\IEEEauthorblockA{\textit{IEO Psychology Division; Univ.}\\
of Milan, Italy\\
ketti.mazzocco@ieo.it}
\and
\IEEEauthorblockN{Anastasia Constantinidou}
\IEEEauthorblockA{\textit{BOCOC}\\
Nicosia, Cyprus\\
anastasia.constantinidou\\
\textit{@bococ.org.cy}
}
\and
\IEEEauthorblockN{Konstantinos Marias}
\IEEEauthorblockA{\textit{ECE Dept. HMU}\\
\textit{FORTH-ICS}\\
Heraklion, Greece\\
kmarias@ics.forth.gr}

\and
\IEEEauthorblockN{Dimitrios I. Fotiadis,}
\IEEEauthorblockA{\textit{Fellow, IEEE, FORTH-BRI}\\
Ioannina, Greece\\
fotiadis@uoi.gr}
\and
\IEEEauthorblockN{Manolis Tsiknakis}
\IEEEauthorblockA{\textit{ECE Dept. HMU}\\ 
\textit{FORTH-ICS}\\
Heraklion, Greece\\
tsiknaki@ics.forth.gr}
}

\maketitle

\begin{abstract}
Frailty and functional decline strongly influence treatment tolerance and outcomes in older patients with cancer, yet assessment is typically limited to infrequent clinic visits. We propose a multimodal wearable framework to estimate frailty-related functional change between visits in elderly breast cancer patients enrolled in the multi-center CARDIOCARE study. Free-living smartwatch physical activity and sleep features are combined with ECG-derived heart-rate-variability (HRV) features from a chest strap and organized into patient--horizon bags aligned to month 3 (M3) and month 6 (M6) follow-ups. Our innovation is an attention-based MIL formulation that fuses irregular, multimodal wearable instances under real-world missingness and weak supervision. An attention-based multi-instance learning (MIL) model with modality-specific MLP encoders (embedding dimension $D{=}128$) aggregates variable-length and partially missing longitudinal instances to predict discretized change-from-baseline classes (worsened/stable/improved) for FACIT-F and handgrip strength. Under subject-independent leave-one-subject-out (LOSO) evaluation, the full multimodal model achieved balanced accuracy/F1 of $0.68{\pm}0.08/0.67{\pm}0.09$ at M3 and $0.70{\pm}0.10/0.69{\pm}0.08$ at M6 for handgrip, and $0.59{\pm}0.04/0.58{\pm}0.06$ at M3 and $0.64{\pm}0.05/0.63{\pm}0.07$ at M6 for FACIT-F. Ablation results indicated that smartwatch activity and sleep provide the strongest predictive information for frailty realted functional changes, while HRV contributes complementary information when fused with smartwatch streams. 
\newline

\textbf{Clinical Relevance—}Wearable-based frailty supports continuous monitoring of deconditioning and recovery between visits, enabling earlier triage and more targeted supportive interventions in cardio-oncology care.
\end{abstract}


\section{Introduction}

Frailty is a common and clinically consequential syndrome in older adults, reflecting reduced physiological reserve and limited ability to recover from stressors \cite{Clegg2013}. In oncology, this vulnerability often appears in practical ways. A patient may rapidly decondition during treatment, struggle with symptom burden, and ultimately require dose reductions, delays, or unplanned care \cite{Hurria2019JAGS, Kalsi2014BJC,Mao2025npjDM}.

In cardio-oncology, the clinical implications of frailty are amplified. Functional decline may evolve alongside cardiovascular risk, subclinical cardiac dysfunction, and treatment related cardiotoxicity. Symptoms such as fatigue, reduced exercise tolerance, and activity limitation may reflect overlapping contributions from cancer, cardiovascular disease, and autonomic dysregulation. As a result, frailty trajectories in this population influence not only oncologic outcomes but also cardiovascular vulnerability and recovery potential during therapy \cite{Coviello2018}. 

Frailty matters because it influences decisions throughout the care pathway, from intensity and supportive measures of treatment to rehabilitation planning and frequency of follow-up \cite{Wildiers2014SIOG}. Despite its importance, frailty is usually captured intermittently through clinic-based assessments (e.g., geriatric evaluations, performance status, functional tests) which are time-consuming and rarely repeated at a cadence that matches real-world change \cite{Owusu2014CGA}. For many older patients, physical function can drift over weeks due to fatigue, sleep disruption, reduced activity, intercurrent illness, or treatment side effects \cite{Wong2018PFtraj}. When measurement is sparse, early warning signs of decline may be missed, and recovery trajectories outside the clinic may go undocumented.

Wearable sensors provide a practical way to observe behavior and physiology repeatedly in free-living conditions. Activity patterns, rest--activity rhythms, sleep behavior, and cardiac features are closely related to frailty-relevant domains such as endurance, mobility, and recovery \cite{Razjouyan2018SensorsFrailty}. The challenge is to turn noisy, incomplete, and heterogeneous streams into a clinically useful frailty estimate. In real deployments, adherence varies, missingness is common, and frailty is not a single physiological quantity, it is a multi-domain construct that manifests through shifts in behavior and function \cite{Daniore2024DACIA}. This makes supervision inherently weak: clinical labels are patient-level and sparse, while wearable data are dense and fragmented. Recent clinical reviews in elderly breast cancer emphasize that cardiotoxicity risk is shaped by multimorbidity and frailty and argue for closer, more systematic surveillance, yet this is difficult to achieve with sparse clinic-based assessments alone \cite{Keramida}.

Against this backdrop, we study frailty-related functional change in an elderly, multi-center cardio-oncology cohort as part of the EU-funded clinical study CARDIOCARE (Grant Agreement No 945175) \cite{CARDIOCARE}. We focus on follow-up timepoints where clinical measurements are available, while wearable signals are recorded longitudinally in free-living conditions. Specifically, we use change-from-baseline targets derived from the fourth version of Functional Assessment of Chronic Illness Therapy (FACIT) \cite{Cella1997FACITManual} and handgrip dynamometer at M3 and M6, treating each follow-up label as supervision for the surrounding wearable evidence.

To address this weak-supervision regime, we propose a multimodal framework that combines modality-specific encoders with attention-based multi-instance learning (MIL) \cite{Ilse2018MIL} to infer frailty-related change from wearable-derived features. Each patient--timepoint is modeled as a variable-length bag of instances drawn from physical activity, sleep, and ECG-derived Heart Rate Variability (HRV), allowing the model to aggregate partially missing evidence without heuristic averaging. The attention mechanism provides a transparent instance-weighting signal that can be inspected to understand which periods contribute most to the final prediction. By targeting an under-studied, clinically meaningful cardio-oncology population, this work aims to advance scalable, non-invasive tools for monitoring deconditioning and recovery between clinic visits, supporting timely triage and individualized supportive care.

\section{Related Work}

Wearable sensors have emerged as practical tools for objectively gauging frailty in older adults, particularly through free-living activity and gait signals used as digital biomarkers \cite{REF_GENERAL_WEARABLE_FRAILTY}. Park \textit{et al.} showed that a short free-living recording with a pendant accelerometer (48~h) can discriminate robust vs.\ non-robust phenotypes using a compact feature set (e.g., time standing/walking, cadence, and longest walking bout), achieving $\approx$75--80\% accuracy against Fried frailty criteria \cite{Park202148h}. Using an instrumented 5-repetition sit-to-stand (5$\times$STS) task, Park \textit{et al.} identified three motion-derived features (hip angular velocity and vertical power statistics) that captured frailty phenotypes and enabled classification with AUC $\approx$0.85 (sensitivity 83\%, specificity 71\%) \cite{Park2021SensorsSTS}. Jung \textit{et al.} showed that wearable gait sensing can support multi-class frailty stratification: an LSTM-based analysis of foot-sensor gyroscope sequences from a 7~m walk achieved strong three-class discrimination (F$_1 \approx 0.93$) when combined with a random-forest classifier \cite{REF_JUNG_LSTM_GAIT}. Fan \textit{et al.} further reported that integrating wearable 6-minute walk test features with clinical variables improved frailty prediction (AUC $\approx$0.93), with gait speed, step length, and total distance among the most informative predictors \cite{REF_FAN_6MWT}. Finally, wrist-worn accelerometry has also been shown to identify non-robust older adults with high sensitivity and specificity, suggesting that convenient device placements can be sufficient for scalable screening in practice \cite{REF_WRIST_VS_LOWERBACK}.

Recently wearable-based frailty assessment in older adults with cancer, building on advancements in geriatric populations, where frailty directly impacts treatment decisions and outcomes \cite{REF_GERONC_CONTEXT}. A recent systematic review indicated that wearable sensing in geriatric oncology has predominantly concentrated on \emph{free-living physical activity} as a surrogate for recovery and functional resilience (e.g., postoperative monitoring or during chemotherapy), and it suggested that continuous monitoring is viable and correlates with clinical assessment metrics \cite{REF_DUIN_REVIEW}. In a pilot study of cancer survivors aged $\geq 65$ years, higher step counts, faster real-world gait, and less fragmented walking over a 4-week period were associated with better patient-reported and performance-based physical function (Pearson $r \approx 0.6$--$0.7$), while peak daily cadence related to fall risk, suggesting wearable gait metrics can flag high-risk individuals \cite{REF_LOW_PILOT}. Beyond association studies, wearables have also been explored for early risk stratification: in older patients receiving chemotherapy, pendant accelerometry collected for 14 days after the first cycle was used to derive frailty-relevant features (e.g., cadence, step count, sit-to-stand transitions, and sustained walking bouts) and a ``Chemotherapy Resilience Index'' that differentiated patients who later experienced treatment-limiting toxicities or unplanned acute care, with large effect sizes when combined with baseline performance status \cite{REF_CAY_CRI}. Overall, these studies indicate that continuous wearable monitoring is both acceptable and operationally feasible in this setting.

Despite growing interest in wearable-based frailty assessment, limited work has focused on cardio-oncology populations, where cardiovascular comorbidity, cardiotoxic treatments, and cardioprotective medications may alter both activity patterns and autonomic markers such as HRV. This gap highlights the need for population-specific validation and interpretation frameworks.

\section{Methodology}

\subsection{Dataset}
The proposed framework was assessed using longitudinal data from participants enrolled in the CARDIOCARE clinical study. The cohort comprises older women with breast cancer, over sixty years of age, who were monitored due to elevated risk of cancer therapy–related cardiotoxicity. Recruitment and data acquisition were conducted across six clinical sites: Bank of Cyprus Oncology Center (BOCOC, No. ΕΕΒΚ/ΕΠ/2022/58), Karolinska Universitetssjukhuset (KSBC, No. 2023-0062-01-413152),  University of Ioannina (UOI, No. 25/23-11/2022), National and Kapodistrian University of Athens (NKUA, No. 456/14-10-2022, ΕΒΔ-683/22-11-2022), Onkoloski Institut Ljubljana (IOL, No. ERIDEK-0038/2023), and Instituto Europeo Di Oncologia SRL (IEO, No. R1754/22-IEO 1874). All participants provided written informed consent before inclusion, and ethical approval was obtained locally at each participating center through the corresponding Institutional Review Boards and Ethics Committees.

Continuous, multimodal sensing was performed through a combination of wearable and clinical-grade devices. Daily-life physical activity and sleep were recorded using the Garmin Venu SQ smartwatch \cite{GarminVenuSQ}, with a minimum data availability criterion of two days per week per participant. Complementary cardiac monitoring was obtained via the Polar H10 chest (Fs = 130Hz) strap \cite{PolarH10}, which provided biweekly ECG recordings of at least 30 minutes during the first six months after enrolment.

Frailty-related functional status was characterized using both patient-reported and objective clinical measures. Fatigue-related quality of life was captured with the FACIT-F version 4, muscular strength was quantified through handgrip dynamometer. These outcomes were collected at baseline (BL) and at follow-up visits at month 3 (M3) and month 6 (M6). For modeling, we used change-from-baseline at M3 and M6 as the prediction targets, representing longitudinal functional deterioration or recovery between visits.

The overall monitoring protocol, including device selection, recording cadence, and follow-up timepoints, was specified by the project’s specialized oncologists to meet CARDIOCARE’s clinical surveillance objectives.

For each endpoint, we compute deltas:
\begin{equation}
\Delta_{h}^{(t)} = y_{h}^{(t)}-y_{\mathrm{BL}}^{(t)}, 
\qquad h\in\{\mathrm{M3},\mathrm{M6}\},
\end{equation}
where $t$ denotes the task (FACIT-F or handgrip). We discretize $\Delta_h^{(t)}$ into three classes using a margin $r_t$:
\begin{equation}
c(\Delta;r_t)=
\begin{cases}
0, & \Delta \le -r_t \quad (\text{worsened}),\\
1, & -r_t < \Delta < r_t \quad (\text{stable}),\\
2, & \Delta \ge r_t \quad (\text{improved}).
\end{cases}
\end{equation}

We set $r_{\mathrm{FACIT}}=5$, reflecting commonly reported clinically meaningful change thresholds for FACIT-Fatigue in the 3--5 point range and adopting a conservative discretization margin \cite{Cella2025FACITCAD}. We further set $r_{\mathrm{HG}}=2$\,kg, since small fluctuations in handgrip strength may reflect measurement variability, with minimal detectable change estimates in older adults on the order of 2--3\,kg \cite{Sawaya2021MDC}.

 The final number of patients included for each endpoint (FACIT-F and handgrip), along with the resulting class distributions per task and follow-up horizon, are shown in Table~\ref{tab:class_dist}.

\begin{table}[!h]
\caption{Class counts (0=worsened, 1=stable, 2=improved).}
\label{tab:class_dist}
\centering
\footnotesize
\setlength{\tabcolsep}{5pt}
\renewcommand{\arraystretch}{1.05}
\begin{tabular}{lrrrr}
\toprule
\textbf{Task} & \textbf{0} & \textbf{1} & \textbf{2} & \textbf{Total} \\
\midrule
FACIT $\Delta$ M3 & 132 & 252 &  78 & 462 \\
FACIT $\Delta$ M6 & 120 & 243 &  99 & 462 \\
Handgrip $\Delta$ M3    &  75 & 124 &  34 & 233 \\
Handgrip $\Delta$ M6    &  77 &  98 &  58 & 233 \\
\bottomrule
\end{tabular}
\end{table}

\subsection{Preprocessing}\label{sec:preprocessing}
We consider three tabular wearable modalities: smartwatch physical activity (\emph{phys}), smartwatch sleep (\emph{sleep}), and ECG-derived heart-rate-variability (\emph{hrv}). Each modality provides repeated observations over time for each patient; each observation date is treated as one \emph{instance}. For patient $p$, let $b_p$ denote the baseline date. For an instance recorded at calendar date $d$, we define the days-from-baseline:
\begin{equation}
\tau(d,p) = (d-b_p)\ \text{days}.
\end{equation}
Instances are aligned to follow-up horizons $h\in\{\mathrm{M3},\mathrm{M6}\}$ using fixed day windows around each scheduled follow-up. Specifically, for each horizon $h$ we define an interval $[t_h^{-},t_h^{+}]$ (in days from baseline) and assign:
\begin{equation}
\mathcal{H}(\tau)=
\begin{cases}
\mathrm{M3}, & t_{\mathrm{M3}}^{-}\le \tau \le t_{\mathrm{M3}}^{+},\\
\mathrm{M6}, & t_{\mathrm{M6}}^{-}\le \tau \le t_{\mathrm{M6}}^{+},\\
\varnothing, & \text{otherwise},
\end{cases}
\label{eq:horizon_rule}
\end{equation}
where $\varnothing$ denotes that the instance is excluded from subsequent analysis. Within each modality, instances are ordered chronologically by their observation date.

For the ECG modality, raw Polar recordings are converted to tabular HRV features using NeuroKit2 \cite{neurokit}. Briefly, R-peaks are detected from the ECG signal and converted to an NN-interval (inter-beat interval) series, from which standard time-domain, frequency-domain, and non-linear HRV indices are computed. When multiple HRV segments are available for a given recording date, segment-level HRV features are aggregated to obtain a single feature vector for that date (e.g., using robust statistics such as the median), yielding one HRV instance per observation date.

To prevent information leakage under patient-level cross-validation, preprocessing steps which estimate parameters are performed using training patients only, and are then applied unchanged to validation and test patients within the same fold. For each modality $m\in\{\mathrm{phys},\mathrm{sleep},\mathrm{hrv}\}$ and feature index $f$, we estimate training-fold statistics $(\mu_{m,f},\sigma_{m,f})$. Missing entries are imputed using $\mu_{m,f}$, and features are standardized via: \begin{equation} \tilde{x}_{m,f}=\frac{x_{m,f}-\mu_{m,f}}{\sigma_{m,f}+\varepsilon}, \label{eq:zscore} \end{equation} where $\varepsilon$ is a small constant to ensure numerical stability.

\subsection{Bag construction}\label{sec:bag_construction}
For each patient $p$ and follow-up horizon: $h\in\{\mathrm{M3},\mathrm{M6}\}$, we construct a multimodal bag
\begin{equation}
\mathcal{B}_{p,h}=\Big(\mathbf{X}^{\mathrm{phys}}_{p,h},\ \mathbf{X}^{\mathrm{sleep}}_{p,h},\ \mathbf{X}^{\mathrm{hrv}}_{p,h}\Big),
\end{equation}
where each modality-specific matrix stacks instances (rows) and features (columns):
\begin{equation}
\mathbf{X}^{m}_{p,h}\in\mathbb{R}^{N^{m}_{p,h}\times F_m},\qquad m\in\{\mathrm{phys},\mathrm{sleep},\mathrm{hrv}\}.
\end{equation}
Here, $N^{m}_{p,h}$ denotes the number of available instances for modality $m$ whose relative time satisfies $\mathcal{H}(\tau)=h$, and $F_m$ denotes the feature dimensionality of modality $m$. Consequently, bags are variable-length across both patients and modalities.

To characterize sensing density and modality imbalance, we report the number of instances per bag stratified by task, horizon, and class label in Table~\ref{tab:instance_stats}. These values are computed after preprocessing, and horizon assignment.

\begin{table}[!h]
\caption{Number of available instances (in thousands, ``k'') by modality, endpoint, follow-up horizon, and delta class (stable/worsened/improved).}
\label{tab:instance_stats}
\centering
\footnotesize
\setlength{\tabcolsep}{5pt}
\renewcommand{\arraystretch}{1.00}
\begin{tabular}{lcccccc}
\toprule
\multirow{2}{*}{\textbf{Modality}} &
\multicolumn{2}{c}{\textbf{Stable}} &
\multicolumn{2}{c}{\textbf{Worsened}} &
\multicolumn{2}{c}{\textbf{Improved}} \\
\cmidrule(lr){2-3}\cmidrule(lr){4-5}\cmidrule(lr){6-7}
& \textbf{M3} & \textbf{M6} & \textbf{M3} & \textbf{M6} & \textbf{M3} & \textbf{M6} \\
\midrule
Garmin Handgrip        & 20.8k & 17.6k & 11.5k & 15.3k &  8.1k & 11.2k \\
Polar Belt Handgrip    &  5.5k &  3.8k &  7.6k &  4.1k &  1.9k &  3.3k \\
Garmin FACIT     & 40.4k & 41.9k & 21.7k & 18.6k & 13.9k & 17.6k \\
Polar FACIT      & 16.6k &  7.9k &  9.7k &  4.5k &  2.9k &  4.0k \\
\bottomrule
\end{tabular}
\end{table}

\subsection{Model architecture and training objective}\label{sec:model}
\begin{figure*}[!htbp]
\centering
\includegraphics[width=0.95\linewidth]{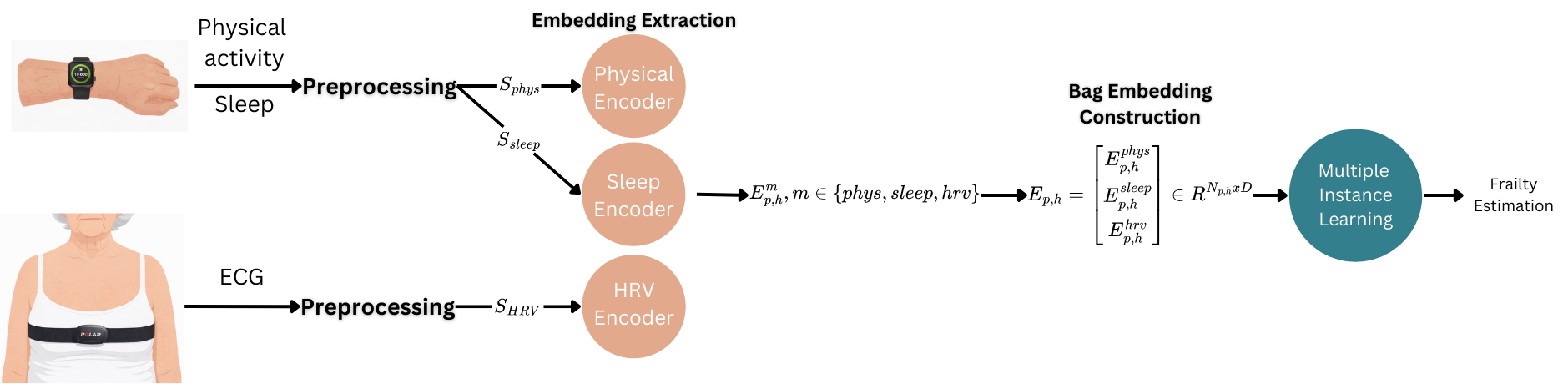}
\caption{A graphical illustration of the proposed pipeline; the initial two images (human icons) were generated using Google Gemini \cite{Gemini} (synthetic images).}
\label{pipeline}
\end{figure*}

In line with a MIL formulation, each patient--horizon pair $(p,h)$ constitutes a \emph{bag} $\mathcal{B}_{p,h}$ comprising a variable number of wearable \emph{instances} across modalities, while supervision is provided only at the bag level through the clinical label $y_{p,h}$. Given the variable-length bag $\mathcal{B}_{p,h}$, the objective is to predict the corresponding 3-class label $y_{p,h}\in\{0,1,2\}$. An overview of the full pipeline is provided in Fig.~\ref{pipeline}.

For each modality $m\in\{\mathrm{phys},\mathrm{sleep},\mathrm{hrv}\}$, an MLP encoder maps each preprocessed instance $\tilde{\mathbf{x}}^{m}_{i}\in\mathbb{R}^{F_m}$ to a shared embedding space:
\begin{equation}
\mathbf{e}^{m}_{i} = f_m(\tilde{\mathbf{x}}^{m}_{i}) \in \mathbb{R}^{D}, \qquad D=128.
\end{equation}
Stacking the instance embeddings yields a modality-specific embedding matrix:
\begin{equation}
\mathbf{E}^{m}_{p,h}
=
\begin{bmatrix}
(\mathbf{e}^{m}_{1})^\top\\
\vdots\\
(\mathbf{e}^{m}_{N^{m}_{p,h}})^\top
\end{bmatrix}
\in \mathbb{R}^{N^{m}_{p,h}\times D}.
\label{eq:Em_def}
\end{equation}

The modality-specific embedding matrices are concatenated to form the bag-level instance embedding matrix:
\begin{equation}
\mathbf{E}_{p,h}
=
\begin{bmatrix}
\mathbf{E}^{\mathrm{phys}}_{p,h}\\
\mathbf{E}^{\mathrm{sleep}}_{p,h}\\
\mathbf{E}^{\mathrm{hrv}}_{p,h}
\end{bmatrix}
\in \mathbb{R}^{N_{p,h}\times D},
\qquad
N_{p,h}=\sum_{m} N^{m}_{p,h}.
\label{eq:Ebag_def}
\end{equation}
Since subsequent pooling is permutation-invariant, the row ordering of $\mathbf{E}_{p,h}$ is immaterial. To preserve modality identity after concatenation, we add a learned modality embedding $\mathbf{v}_m\in\mathbb{R}^{D}$ to each row corresponding to modality $m$. Denoting by $m(j)$ the modality of row $j$, we write:
\begin{equation}
\tilde{\mathbf{e}}_j = \mathbf{e}_j + \mathbf{v}_{m(j)},\qquad j=1,\dots,N_{p,h},
\end{equation}
and let $\tilde{\mathbf{E}}_{p,h}\in\mathbb{R}^{N_{p,h}\times D}$ be the matrix obtained by applying this operation row-wise to $\mathbf{E}_{p,h}$.

Each tagged instance embedding is transformed by a projector $\phi(\cdot)$ into $\mathbf{h}_j=\phi(\tilde{\mathbf{e}}_j)$. An attention network $a(\cdot)$ produces instance scores $\ell_j=a(\mathbf{h}_j)$, which are normalized within the bag:
\begin{equation}
\alpha_j=\frac{\exp(\ell_j)}{\sum_{k=1}^{N_{p,h}}\exp(\ell_k)}.
\end{equation}
The bag representation is computed as an attention-weighted sum,
\begin{equation}
\mathbf{z}_{p,h}=\sum_{j=1}^{N_{p,h}}\alpha_j\,\mathbf{h}_j.
\label{eq:z_bag}
\end{equation}
This attention-based pooling implements the MIL assumption by learning a permutation-invariant aggregation of instance embeddings and assigning higher weight to instances which are most informative for predicting the bag label.

A classifier head produces bag-level logits $\hat{\mathbf{y}}_{p,h}=g(\mathbf{z}_{p,h})\in\mathbb{R}^{3}$. The training objective is the (optionally class-weighted) cross-entropy loss:
\begin{equation}
\mathcal{L} = - \sum_{c=0}^{2} w_c\,\mathbb{I}[y_{p,h}=c]\ \log\Big(\mathrm{softmax}(\hat{\mathbf{y}}_{p,h})_c\Big),
\end{equation}
where $w_c$ may be computed from the training data to mitigate class imbalance.

\section{Experiments and Results}

\subsection{Evaluation protocol}\label{sec:eval_protocol}
Model performance was assessed under subject-independent evaluation to reflect deployment on previously unseen patients. Specifically, we employed a patient-level leave-one-subject-out (LOSO) protocol, in which all samples (bags) from a held-out patient were excluded from training and used only for testing in that iteration. Within each LOSO iteration, the remaining patients were used to train the model, with a patient-wise split of the training pool into training and validation subsets (80\%-20\%) for model selection and early stopping. All preprocessing operations which estimate parameters (e.g., imputation statistics and feature standardization) were fitted exclusively on the training patients of the current iteration and then applied unchanged to the corresponding validation and test patients to avoid information leakage. Performance is reported using balanced accuracy and F1 score, summarized as mean$\pm$standard deviation across LOSO iterations.

\subsection{Experimental setup}\label{sec:exp_setup}
We trained an MIL classifier that operates on multimodal bags constructed per patient and follow-up horizon. Each bag comprises a variable number of instances from smartwatch physical activity features, smartwatch sleep features, and ECG-derived HRV features. Modality-specific MLP encoders map each instance into a shared embedding space, after which all instance embeddings are concatenated and aggregated using attention pooling to form a fixed-dimensional bag representation. A linear classification head predicts the 3-class outcome (worsened/stable/improved) for each patient--horizon pair. Training minimizes the cross-entropy loss. We ensured reproducibility by adopting a deterministic training setup, fixing all random seeds and enforcing deterministic execution in the learning framework and data-loading pipeline for all experiments. All hyperparameters were fixed across
experiments and ablations. Model size and computational cost are modest (\textbf{0.307}M parameters; \textbf{0.479}G FLOPs).

\subsection{Results}\label{sec:main_results}
Table~\ref{tab:main_results} summarizes the overall performance of the full multimodal model for both endpoints at months 3 and 6. For handgrip strength, the model achieved a balanced accuracy of $0.68\pm0.08$ at M3 and $0.70\pm0.10$ at M6, with corresponding F1 scores of $0.67\pm0.09$ and $0.69\pm0.08$. For FACIT-F, performance was lower at M3 (balanced accuracy $0.59\pm0.04$, F1 $0.58\pm0.06$) and improved at M6 (balanced accuracy $0.64\pm0.05$, F1 $0.63\pm0.07$), indicating more reliable discrimination of longitudinal functional change at the later follow-up.

\begin{table}[!h]
\caption{Overall results (mean$\pm$std across LOSO iterations) for the full multimodal MIL model.}
\label{tab:main_results}
\centering
\setlength{\tabcolsep}{8pt}
\renewcommand{\arraystretch}{1.05}
\begin{tabular}{llcc}
\toprule
\textbf{Endpoint} & \textbf{Metric} & \textbf{M3} & \textbf{M6} \\
\midrule
\multirow{2}{*}{Hg}
& Balanced Accuracy & $0.68\pm0.08$ & $\mathbf{0.70\pm0.10}$ \\
& F1 score          & $0.67\pm0.09$ & $\mathbf{0.69\pm0.08}$ \\
\midrule
\multirow{2}{*}{FACIT}
& Balanced Accuracy & $0.59\pm0.04$ & $\mathbf{0.64\pm0.05}$ \\
& F1 score          & $0.58\pm0.06$ & $\mathbf{0.63\pm0.07}$ \\
\bottomrule
\end{tabular}
\end{table}

\subsection{Ablation study}\label{sec:ablation_results}
We further quantified the contribution of each sensing modality by training the same MIL architecture using only modality pairs: physical activity+sleep (P+S), physical activity+ECG/HRV (P+E), and sleep+ECG/HRV (S+E). Results are reported in Table~\ref{tab:ablation}. Across both endpoints and horizons, P+S consistently outperformed the other two pairs, suggesting that the smartwatch-derived streams provide the strongest standalone signal for discriminating functional change. In contrast, S+E produced the lowest performance, particularly at M3, indicating that activity features are critical for robust classification. Importantly, the full multimodal configuration exceeded all pairwise settings for both endpoints, with the largest gains observed for FACIT-F, consistent with complementary information contributed by ECG-derived HRV when combined with smartwatch data.

\begin{table}[!htbp]
\caption{Ablation results (mean$\pm$std across LOSO iterations) using modality pairs.}
\label{tab:ablation}
\centering

\setlength{\tabcolsep}{5pt}
\renewcommand{\arraystretch}{1.05}
\begin{tabular}{llcccc}
\toprule
\textbf{Split} & \textbf{Endpoint} &
\shortstack{\textbf{BalAcc}\\\textbf{M3}} &
\shortstack{\textbf{F1}\\\textbf{M3}} &
\shortstack{\textbf{BalAcc}\\\textbf{M6}} &
\shortstack{\textbf{F1}\\\textbf{M6}} \\
\midrule
\multirow{2}{*}{P+S}
 & Hg    & \textbf{0.65$\pm$0.04} & \textbf{0.63$\pm$0.02} & \textbf{0.67$\pm$0.05} & \textbf{0.65$\pm$0.03} \\
 & FACIT & \textbf{0.52$\pm$0.07} & \textbf{0.50$\pm$0.08} & \textbf{0.57$\pm$0.06} & \textbf{0.55$\pm$0.04} \\
\midrule
\multirow{2}{*}{P+E}
 & Hg    & 0.60$\pm$0.04 & 0.61$\pm$0.03 & 0.62$\pm$0.05 & 0.62$\pm$0.04 \\
 & FACIT & 0.50$\pm$0.02 & 0.49$\pm$0.01 & 0.51$\pm$0.03 & 0.51$\pm$0.04 \\
\midrule
\multirow{2}{*}{S+E}
 & Hg    & 0.50$\pm$0.03 & 0.49$\pm$0.05 & 0.54$\pm$0.05 & 0.55$\pm$0.03 \\
 & FACIT & 0.46$\pm$0.05 & 0.46$\pm$0.01 & 0.48$\pm$0.06 & 0.49$\pm$0.05 \\
\bottomrule
\end{tabular}
\end{table}

\section{Discussion}

In this study, we examined whether longitudinal, free-living wearable data can support estimation of frailty-related functional change between clinic visits in an elderly, multi-center cardio-oncology cohort. Using discretized change-from-baseline targets at M3 and M6 derived from FACIT-F and handgrip strength, the proposed attention-based MIL model achieved moderate subject-independent performance under a LOSO protocol. Performance was consistently higher for handgrip than for FACIT-F and improved at M6 for both endpoints. For handgrip, balanced accuracy/F1 reached $0.68\pm0.08$/$0.67\pm0.09$ at M3 and $0.70\pm0.10$/$0.69\pm0.08$ at M6. For FACIT-F, performance increased from $0.59\pm0.04$/$0.58\pm0.06$ at M3 to $0.64\pm0.05$/$0.63\pm0.07$ at M6. Overall, these findings suggest that multimodal wearable-derived behavioral and physiological signals capture discriminative information about functional trajectories, with clearer separation at the later follow-up.

Variations among endpoints are clinically plausible. Handgrip strength is an objective measure of muscular function and is typically less influenced by transient contextual and psychosocial factors at the time of assessment. In contrast, FACIT-F reflects perceived fatigue and broader quality-of-life dimensions, which may vary with symptoms, mood, treatment side effects, and environmental conditions which are only partially represented by activity, sleep, or short ECG-derived HRV segments. Moreover, discretizing continuous deltas into three classes introduces ambiguity near decision thresholds, which can effectively increase label noise, particularly for subjective outcomes. The stronger discrimination at M6 may indicate that longer term changes yield a more stable wearable signature as sustained deconditioning or recovery becomes more apparent than short-lived fluctuations.

The ablation analysis further clarifies the contribution of each sensing stream. Across endpoints and horizons, the physical-activity plus sleep pairing (P+S) consistently outperformed the other modality pairs, whereas configurations excluding activity were weaker. This aligns with the close relationship between free-living mobility, rest--activity rhythms, and functional status, with sleep providing complementary context. By contrast, the sleep+ECG/HRV pairing (S+E) produced the lowest performance, which may reflect the lower recording cadence of ECG sessions and the sensitivity of HRV to medication effects, acute symptoms, and recording conditions. Importantly, the full multimodal configuration exceeded the pairwise settings, with the largest relative gains observed for FACIT-F, suggesting that HRV provides complementary physiological context when fused with smartwatch-derived behavior.

Methodologically, the key innovation is an attention-based, multimodal MIL formulation that performs variable-length fusion across heterogeneous wearable streams under real-world missingness, aligning dense free-living evidence to sparse patient--horizon supervision without heuristic averaging. Attention pooling provides permutation-invariant aggregation and yields instance-level weights that support model inspection and post hoc interpretation. 

In practice, wearable-based detection of a worsened frailty trajectory could enable targeted clinical actions, including expedited cardio-oncology assessment, optimization of cardioprotective therapy, referral to exercise or rehabilitation programs, intensified supportive care (e.g., nutrition and symptom management), and—when clinically appropriate—adjustment of cancer treatment intensity in higher-risk individuals.

This study is limited by weak supervision and discretization-induced label ambiguity, heterogeneous adherence and center effects, and the relatively sparse cadence and context sensitivity of HRV recordings, which together constrain robustness and generalizability.

Several extensions could improve clinical readiness and evaluation. Methodologically, robustness may be strengthened via center-aware normalization or domain adaptation to mitigate site and adherence effects, and by replacing hard binning with ordinal losses and/or continuous delta regression to reduce label ambiguity. For deployment, uncertainty estimation and probability calibration are important, particularly near decision thresholds, and longitudinal multi-task learning across M3/M6 and endpoints (FACIT-F, handgrip) may improve sample efficiency and consistency. Evaluation should move beyond snapshot classification to clinically aligned objectives, including within-patient trend detection, early warning of sustained decline, and associations with downstream outcomes such as treatment tolerance, unplanned care utilization, and cardiotoxicity-related symptoms.

\section{Conclusion}

In this work we introduced a multimodal wearable MIL approach to estimate frailty-related functional change between visits in elderly cardio-oncology, enabling variable-length fusion under real-world missingness. LOSO evaluation shows moderate discrimination of worsened/stable/improved change, with smartwatch streams contributing most and HRV providing complementary information. 

Wearable-based frailty monitoring has the potential to function as an early warning tool within cardio-oncology care pathways, supporting proactive cardiovascular surveillance and individualized supportive care in older patients undergoing cancer therapy, future work will focus on robustness to site effects and clinically actionable longitudinal monitoring.

\vspace{12pt}

\end{document}